\documentclass{article} 
\usepackage{iclr2023_conference,times}
\iclrfinalcopy

\usepackage{amsmath,amsfonts,bm}

\def\eqref#1{equation~\ref{#1}}

\def\1{\bm{1}}

\DeclareMathAlphabet{\mathsfit}{\encodingdefault}{\sfdefault}{m}{sl}
\SetMathAlphabet{\mathsfit}{bold}{\encodingdefault}{\sfdefault}{bx}{n}

\usepackage{hyperref}
\usepackage{url}

\usepackage{booktabs} 
\usepackage{tikz}
\usetikzlibrary{arrows}
\usetikzlibrary{shapes.multipart}
\usetikzlibrary{patterns}
\usepackage{caption}

\usepackage{booktabs,algorithm}
\usepackage{algcompatible}
\usepackage[noend]{algpseudocode}

\usepackage[all]{nowidow}

\usepackage[font=small,labelfont=bf]{caption}
\usepackage[utf8]{inputenc} 
\usepackage[T1]{fontenc}    
\usepackage{hyperref}       
\hypersetup{
    colorlinks=false,
    linkcolor=blue,
    filecolor=magenta,      
    urlcolor=cyan,
    pdfpagemode=FullScreen,
    }
\usepackage{url}            
\usepackage{booktabs}       
\usepackage{amsfonts}       
\usepackage{nicefrac}       
\usepackage{microtype}      
\usepackage{xcolor}         
\usepackage{amsmath}
\usepackage{mathtools}
\usepackage{amsthm}
\usepackage{amsmath}
\usepackage{amssymb}
\usepackage{bbm}
\usepackage{tikz} 
\usepackage{wrapfig}

\usepackage{array}
\newcolumntype{H}{>{\setbox0=\hbox\bgroup}c<{\egroup}@{}}

\usepackage{graphicx}

\usepackage{caption}
\usepackage{subcaption}

\usepackage{booktabs,algorithm, amsmath}
\usepackage{algcompatible}
\usepackage[noend]{algpseudocode}

\newcommand{\simevals}{\texttt{SimEvals}}
\newcommand{\simeval}{\texttt{SimEval}}

\title{A Case Study on Designing Evaluations of ML Explanations with Simulated User Studies}

\begin{document}

\author{Ada Martin\thanks{Work done as an intern at Feedzai}  \& Valerie Chen \\
Department of Computer Science\\
Carnegie Mellon University\\
Pittsburgh, PA 15213, USA \\
\texttt{\{gamartin, vchen2\}@cs.cmu.edu} \\
\And
S\'ergio Jesus \& Pedro Saleiro \\
Feedzai \\
Lisbon, Portugal
}

\date{}

\maketitle

\begin{abstract}
When conducting user studies to ascertain the usefulness of model explanations in aiding human decision-making, it is important to use real-world use cases, data, and users. However, this process can be resource-intensive, allowing only a limited number of explanation methods to be evaluated.  
Simulated user evaluations (\simevals{}), which use machine learning models as a proxy for human users, have been proposed as an intermediate step to select promising explanation methods. 
In this work, we conduct the first \simevals{} on a real-world use case to evaluate whether explanations can better support ML-assisted decision-making in e-commerce fraud detection. 
We study whether \simevals{} can corroborate findings from a user study conducted in this fraud detection context. In particular, we find that \simevals{} suggest that all considered explainers are equally performant, and none beat a baseline without explanations -- this matches the conclusions of the user study. Such correspondences between our results and the original user study provide initial evidence in favor of using \simevals{} before running user studies.
We also explore the use of \simevals{} as a cheap proxy to explore an alternative user study set-up.
We hope that this work motivates further study of when and how \simevals{} should be used to aid in the design of real-world evaluations. 


\end{abstract}


\section{Introduction}

The field of interpretable machine learning has proposed a large and diverse number of techniques to explain model behavior. However, it is difficult to anticipate exactly which explanations may help humans with a particular use case~\citep{chen2022interpretable, davis2020measure}.
There have been calls for more human-centered approaches~\cite{vaughan2021humancentered,liao2021human} to investigate how humans benefit from explanations in specific use cases, particularly through user studies~\citep{doshi2017towards}.
Ideally, these user studies would utilize real users, tasks, and data to maximize the applicability of the study's findings~\citep{amarasinghe2020explainable}. 
Since real-world user studies can be resource-intensive to conduct and thus typically only evaluate a limited number of explanation methods (or explainers), simulated user evaluations (\simevals{}) have been proposed as a way to identify candidate explanation methods for user studies using machine learning models~\citep{chen2022use}. While the original work by~\citet{chen2022use} performed a cursory evaluation of \simevals{}, it is unclear whether this  approach would generalize to real-world use cases of explanations.



In this work, we focus on a real-world decision support use case where professional fraud analysts review e-commerce transactions to determine whether a transaction is fraudulent. 
We conduct the first \simevals{} on a real-world task and data and compare the results to the findings from a user study with real-world users conducted by~\citet{amarasinghe2022importance} as shown in Figure~\ref{fig:summary_1}.  
We instantiate \simevals{} to study whether any of these explanations contained predictive information about whether a transaction was fraudulent and find no statistical difference in \simeval{} performance between the three explanation methods and a baseline \simeval{} without explanations. The results of this \simeval{} trial closely match the findings of~\citet{amarasinghe2022importance}.
Our results suggest that \simevals{} could have helped to select better candidate explainers in the original user study, reducing its cost and improving its chance of locating a successful explainer. They also provide evidence that \simeval{} performance is associated with human performance across different explainers. 

We also explore the use of \simevals{} to cheaply identify an alternative study design beyond the canonical set-up where analysts are provided both the transaction and the explanation as shown in Figure~\ref{fig:summary_2}. Our preliminary findings suggest that a subset of the explainers considered in the original study can be used as a human-centric dimensionality reduction technique (i.e., there is not statistically less signal in only presenting the explanation on its own) to reduce the time cost of processing a full transaction. 
To get an initial signal on the validity of this proposed design, we conduct short interviews with multiple fraud analysts and evaluate whether the information the analysts typically look for in a full transaction is 
present in the explanations used in the alternative study design.

In summary, this work explores two ways to utilize \simevals{} in a real-world context. We believe that our comparative investigation of real and simulated user studies will serve as an example of using \simevals{} more effectively.

\section{Can \simevals{} corroborate findings from 
 real-world user study?}\label{sec:comparison}

\begin{figure}
    \centering
    \includegraphics[scale=0.65]{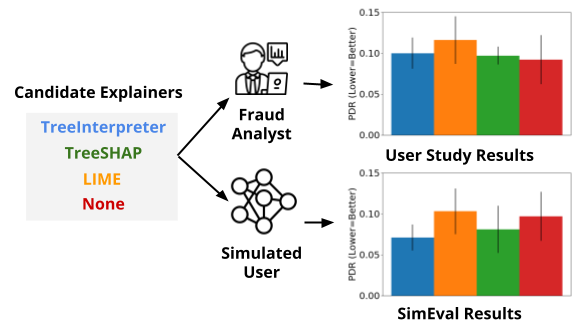}
    \caption{We compare real and simulated user studies of explanations (\simevals{}) for the real-world use case of fraud detection decision support, shown in the top and bottom rows respectively. A user study designed to compare three different study arms, each with a different explanation method, in order of performance. A corresponding \simeval{} study that compares the same three explanation methods but replaces fraud analysts are replaced with ML models. We compare the relative performance of \simeval{} across a variety of candidate explainers against results across comparable arms of~\citet{amarasinghe2022importance} and we find that \simevals{} support the negative result in the user study (i.e., none of the explainers outperform showing the no explanation condition). 
    }
    \label{fig:summary_1}
\end{figure}

While initial results from prior work suggest that \simevals{} can be used to identify promising explanations for user studies, there has been limited evaluation of its utility in real-world contexts. 
We conduct such an evaluation to see whether \simevals{} confirm user study findings in a fraud detection use case.
We first summarize the findings from the original study by~\citet{amarasinghe2022importance} in this use case and then discuss how to instantiate and train a \simeval{} for each explanation method in the study. Figure~\ref{fig:summary_1} summarizes the findings of this section.

\subsection{Prior findings from real-world user study}

The user study by~\citet{amarasinghe2022importance} investigated the real-world application of detecting financial fraud.\footnote{Note that this use case was studied by both~\citet{amarasinghe2022importance} and an earlier study by~\citet{feedzaipaper}. We chose to compare our \simevals{} with findings from the more recent user study because it improved the experimental design in multiple ways to be more representative of the real-world use case.} In the fraud detection use case, a machine learning model $f$ is trained to estimate $\hat{y}$ which is the probability that a given transaction was actually fraud given a piece of transaction data $x$. The user study introduced an explanation $E(x,f)$ of the model $f$ for a given transaction $x$ to the fraud analysts, hypothesizing  
that this additional information could improve decision outcomes and speed. Specifically, each $E(x,f)$ is a sparse vector which contains feature importance for the top-6 highest magnitude importance features for a given transaction $x$ and model $f$ and 0 otherwise. To decide whether $x$ was fraudulent, analysts were given the transaction $x$ (which had 112 features), the explanation $E(x,f)$, and the model score $\hat{y} = f(x)$. In addition, there were two baseline arms in which the analysts were given only $x$ or $(x, \hat{y})$ to predict fraud.

Analysts in the study were shown 500 transactions for each of three different explainers (LIME~\citep{ribeiro2016should}, TreeSHAP~\citep{lundberg2017unified}, TreeInterpreter~\citep{saabas2015}) and for each baseline arm. ~\citet{amarasinghe2022importance} proposed a metric called \emph{Percent Dollar Regret} (PDR) to better reflect operational goals. PDR measures the amount of revenue lost due to incorrect decisions relative to what would be realized if all the reviewed transactions were perfectly classified:
\begin{equation}
    PDR =  1-\frac{\textrm{Realized~Revenue}}{\textrm{Possible~Revenue}}
\end{equation}
The more detailed equation is found in~\citet{amarasinghe2022importance}. Given this set-up, the main findings of the experiment were: (1) No explanation improved analyst performance in terms of PDR over the baseline of showing analysts the model score only; (2) There was no statistical difference in analyst performance between the three explanation methods.
In this work, we evaluate \simevals{} for both claims to determine whether there was predictive information in any of the explanations that did not translate to improved analyst performance.

\subsection{Setting up \simevals{} to reflect user study}\label{sec:setup}

\simevals{} are ML models trained to predict the ground truth label (e.g., whether a transaction is fraud) given the same information that would be presented in a user study. 
Specifically, the information in the user study can be represented by the tuple $(x, \hat{y}, E(x,f))$, where the explainer $E(\cdot)$ was either TreeInterpreter, LIME, TreeSHAP, or in the baseline case, no explanation and $\hat{y}$ the predicted probability of fraud by $f$.
Each \simeval{} model corresponds to one candidate explanation method.
Validation set PDR is used to evaluate \simevals{}. 
As noted in~\citet{chen2022use}, \simevals{} do not aim to replicate a user's decision-making process and their results should be interpreted as measures of the predictive power of their given explanations. 


\begin{figure}
\begin{center}
\begin{tikzpicture}
\node[anchor=north] at (0.5, 0) {LIME};
\draw(0,0) rectangle (1,4);
\fill[blue!40!white] (0,2.99) rectangle (1,4);
\fill[red!40!white] (0,0.99) rectangle (1,3);
\fill[color=black!80!white] (0,0) rectangle (1,1);

\node[anchor=north] at (2, 0) {SHAP};
\draw(1.5,0) rectangle (2.5,4);
\fill[red!40!white] (1.5,2.99) rectangle (2.5,4);
\fill[blue!40!white] (1.5,1.99) rectangle (2.5,3);
\fill[red!40!white] (1.5,0.99) rectangle (2.5,2);
\fill[color=black!80!white] (1.5,0) rectangle (2.5,1);

\node[anchor=north] at (3.5, 0) {TreeInt};
\draw(3,0) rectangle (4,4);
\fill[red!40!white] (3,2.99) rectangle (4,4);
\fill[red!40!white] (3,1.99) rectangle (4,3);
\fill[blue!40!white] (3,0.99) rectangle (4,2);
\fill[color=black!80!white] (3,0) rectangle (4,1);

\node[anchor=north] at (5, 0) {No Expl};
\draw(4.5,0) rectangle (5.5,4);
\fill[pattern=north east lines, pattern color=red!40!white] (4.5,0.99) rectangle (5.5,4);
\fill[blue!40!white] (4.5,0) rectangle (5.5,1);

\node[anchor=west] at (6.5, 3.75){Train Data};
\draw (6, 4) rectangle (6.5, 3.5);
\fill[color=red!40!white] (6, 4) rectangle (6.5, 3.5);

\node[anchor=west] at (6.5, 3.25){Train Data (Subsampled)};
\draw (6, 3.5) rectangle (6.5, 3);
\fill[pattern=north east lines, pattern color=red!40!white] (6, 3.5) rectangle (6.5, 3);

\node[anchor=west] at (6.5, 2.75){Validation Data};
\draw (6, 3) rectangle (6.5, 2.5);
\fill[color=blue!40!white] (6, 3) rectangle (6.5, 2.5);

\node[anchor=west] at (6.5, 2.25){Unused Data};
\draw (6, 2.5) rectangle (6.5, 2);
\fill[color=black!80!white] (6, 2.5) rectangle (6.5, 2);

\end{tikzpicture}
    \caption{A diagram illustrating the train/test split used for each \simeval{} experiment. As we did not have explanations available for the 500 transactions used in the 'no explanation' arm of the original user study, we perform the above data split. The above data split ensures that the validation dataset associated with each explanation matches the transactions used in the original user study. It also ensures that each SimEval receives the same dataset size (1000 train, 500 validation).}
    \label{fig:dataset_explanation}

    \end{center}
\end{figure}
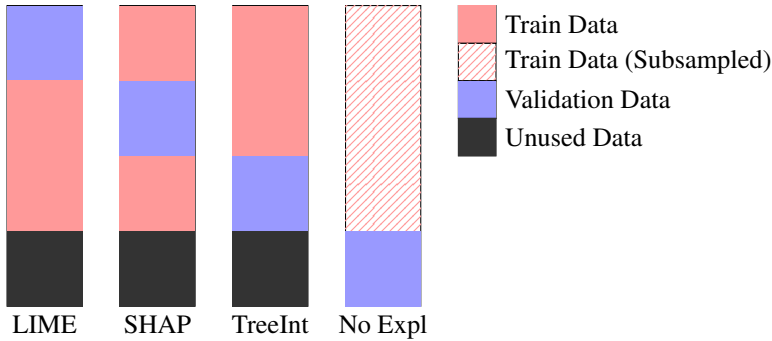

Each \simeval{} was trained and evaluated on $n=1500$ total transactions, which were split into $1000$ train and $500$ validation transactions. The transactions in each split were chosen to match the ones shown to the analysts in the original user study to reduce the impact of the validation set choice on final conclusions. 
Note that this means the validation split is not the same across different explainers because different transactions were shown to the analysts across different arms of the experiment to avoid showing repeated transactions, as shown in Figure~\ref{fig:dataset_explanation}. However, we verified that the different training and validation splits followed roughly the same distribution. 
To select a family of \simeval{} models, we ran a hyperparameter grid search over the parameters in Table~\ref{tab:hyperparams}. We found that the best validation performance was achieved using a Random Forest model with a minimum of 5 samples per leaf node. We use this as the base model family for \simevals{} in the remaining experiments.

\subsection{Checking for parroting}


One unique challenge of decision support use cases like the fraud detection one we consider is the similarity of information provided to users (e.g., the $\hat{y}$, the model prediction for $x$) and the output that the analysts aimed to predict. An effective degenerate strategy emerges for the corresponding \simeval{} model to simply apply a threshold to $\hat{y}$. A \simeval{} model which parrots $\hat{y}$ in this manner will disregard the explainers entirely. When we compare the Original Model results with \simevals{} in Table~\ref{tab:performance}, we find that all explanations improved in PDR when explainers were included, suggesting that the \simevals{} could not have utilized only $\hat{y}$ when making predictions. 



\begin{table}[h!]
\centering
\caption{Performance measured using PDR (lower is better) of the original fraud detection model, \simevals{} and actual analysts across explainers. 
Parentheses contain 90 percent CIs. CIs were obtained by bootstrapping test samples to generate a pivotal confidence interval -- we share CIs rather than standard errors as the CIs are not symmetric. Note that the Original Model column does not depend on the explanations, but the PDR differs due to the different validation split associated with each explainer.
}
\begin{tabular}{l | l | l | l}
& Original Model & \simevals{} & Analyst  \\
\hline
TreeInterpreter & 0.102 (0.075, 0.121)  & 0.071 (0.054, 0.087) & 0.100 (0.081, 0.119)\\
LIME &  0.164 (0.099, 0.226) & 0.103 (0.061, 0.131) & 0.116 (0.087, 0.145)\\
TreeSHAP & 0.109 (0.078, 0.142) & 0.081 (0.052, 0.105) & 0.097 (0.078, 0.116)\\
Model Score & 0.133 (0.082, 0.172) & 0.097 (0.047, 0.127) & 0.092 (0.068, 0.112)\end{tabular}
\label{tab:performance}
\end{table}

\subsection{Comparing \simevals{} to user study findings}

To evaluate whether \simevals{} can corroborate user study findings, we test whether or not the inclusion of explanations generated by any explainer yields higher \simeval{} PDR as compared to a baseline \simeval{} trained without explanations.
We present both the aggregate \simeval{} PDR scores across the validation set for each explainer, which is equivalent to the metric from the user study, as well as a transaction-based analysis where we compare \simeval{} predictions on individual instances with analyst predictions.
A high correspondence on individual transactions would suggest that \simevals{} make decisions in a way that is similar to the analysts.


    \textbf{Comparison of aggregate performance.} In Table~\ref{tab:performance}, we compared \simevals{} to the aggregate analyst performance as found in~\citet{amarasinghe2022importance}. The \simeval{} results show that the overlap in the actual information provided to the analysts across explanations are roughly the same (noting the error bars in the \simevals{} column), which supports the general finding from~\citet{amarasinghe2022importance} that all explainers lead to comparable performance as the baseline (noting the error bars in the Analyst column). Additionally, while not statistically significant, we do observe that both TreeInterpreter and TreeSHAP allowed \emph{both} \simevals{} and analysts to achieve lower PDR compared to LIME.\footnote{Our reproduced analyst PDR for only the model score case differed slightly from the result found in ~\citet{amarasinghe2022importance} (0.092 vs. 0.089). This could be due to different choices in the exclusion of ``warm-up'' samples or a discrepancy in data preprocessing. We suspect that this difference only appeared in the model score case because this data was stored in a separate database than in the other experimental arms.}



\textbf{Comparison on individual transactions.} 
We performed 
an analysis on the association between analyst and \simeval{} predictions on individual transactions. 
Table~\ref{tab:analyst_ROC} shows the ROC AUC when using \simeval{} output as an estimate of the probability that an analyst predicted a given transaction to be fraud. This analysis yielded results which were significantly above 0.5, indicating some association. However, this association was not significantly stronger than when using predictions from the fraud model ($\hat{y}$) as an estimate for analyst predictions directly, implying \simevals{} are may not be as informative at the individual transaction level. 

\begin{table}[h!]
\centering
\caption{The ROC AUC (higher is better) achieved when using either $\hat{y}$ or \simevals{} output to predict the analyst predictions, separated by the type of explainer. We see that \simevals{} and the Original Model are about equally predictive of analyst predictions.
}
\begin{tabular}{lrr}
\toprule
{} &  Original Model &  \simevals{} \\
\midrule
TreeInterpreter &        0.734 &        0.727 \\
LIME            &        0.695 &        0.675 \\
TreeSHAP        &        0.703 &        0.731 \\
\bottomrule
\end{tabular}
\label{tab:analyst_ROC}
\end{table}

\subsection{Discussion \& Limitations}

Since \simevals{} corroborate findings from the user study, we believe the original study could have benefited from running \simevals{} to potentially select better explanation methods before conducting a full user study. 
However, we emphasize that it is not a replacement for running actual user studies. 
The aggregate analysis only provides an estimate of which explainers have the highest performance with no guarantee of how large the difference will actually be in a user study. 
In particular, we might expect more divergence between human and \simeval{} behavior for a few potential reasons: outside domain knowledge is especially important and the analysts lack the time to carefully examine each piece of information as a model would. This divergence is reflected in the modest association between human and \simeval{} predictions shown in Table~\ref{tab:analyst_ROC}.
Although \simevals{} are intended to find predictive information in explanations, it is possible that our choice of base model family or learning procedure may fail to extract this information.  We also note that only the aggregate analysis is possible to conduct before  running a user study, whereas the transaction-level comparison is only possible after a user study has already been run. 


\section{Using \simevals{} to guide new hypotheses} \label{sec:qualitative}

\begin{figure}
    \centering
    \includegraphics[scale=0.54]{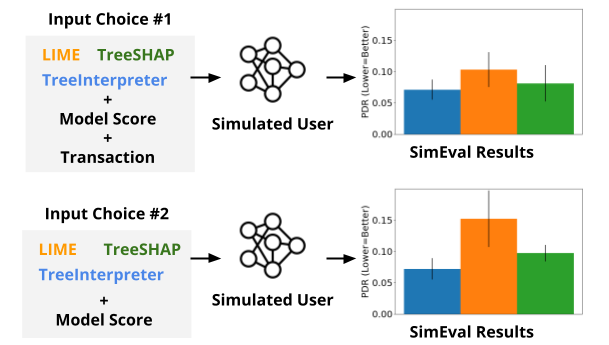}
    \caption{We use \simevals{} to explore a new study design which would only provide the explanation without the input transaction and find that for two explanations there is minimal difference in performance between the two \simeval{} variants. We also conduct short interviews with fraud analysts to get preliminary signal on this study design.
    }
    \label{fig:summary_2}
\end{figure}


Once \simevals{} are set up for a use case, it is easy to vary the parameters of the set-up, which include the choice of inputs. In particular, we explore whether \simevals{} would perform as well when $x$ was excluded from the input (i.e., we train \simevals{} models in the same way as described in Section~\ref{sec:setup} using $(\hat{y}, E(x,f))$ as the inputs). Since $\hat{y}$ is trained on $x$, we hypothesized that it may be redundant to include $x$. When comparing the canonical \simeval{} models with ones that \textit{exclude $x$}, we find that two of the three explainers (TreeInterpreter and TreeSHAP) had only a small performance boost while LIME's performance gap was statistically significant as shown in Table~\ref{tab:performancecompare}. This finding suggests providing analysts only $(\hat{y}, E(x,f))$ for TreeInterpreter and TreeSHAP explainers could reduce the time an analyst spends on each transaction with minimal loss in information content. We summarize this line of analysis in Figure~\ref{fig:summary_2}.
Recall that the visualization of each explanation only shows the top 6 features out of 112.

\begin{table}[h!]
\centering
\caption{Performance of canonical \simevals{} compared with a variant which excludes $x$ as part of the input, measured in PDR (lower is better). Parentheses contain 90 percent CIs. CIs for simulated users were obtained by bootstrapping test samples to generate a pivotal confidence interval.
}
\begin{tabular}{l | l | l }
& \simevals{} excluding $x$ & \simevals{}  \\
\hline
TreeInterpreter  & 0.072 (0.049, 0.089) & 0.071 (0.054, 0.087) \\
LIME  & 0.152 (0.124, 0.197) & 0.103 (0.061, 0.131)\\
TreeSHAP  & 0.097 (0.076, 0.120) & 0.081 (0.052, 0.105)
\end{tabular}
\label{tab:performancecompare}
\end{table}

To evaluate whether a set-up in which analysts are only shown $(\hat{y}, E(x,f))$ may be justified, we perform some initial verification with the analysts. In particular, we investigate whether the features used in $E(x,f)$ across different transactions $x$ have reasonable alignment with features that analysts think are important because it may be unnatural for the analysts to see only explanations consisting of features which they would not typically use.
 


\subsection{Identifying which features are important to the analysts}\label{sec:featimportance}


To obtain analyst's perceived feature importances, we conducted brief interviews with analysts from the original user study by~\citet{amarasinghe2022importance}.
In the interview, we asked analysts to fill out a spreadsheet, where they were asked to rank the importance of each feature in a transaction. They were asked to do this with the context which contained a row for each feature in a transaction. There was also a column for each potential ``transaction reason", where each reason could be considered as a fraudulent or a legitimate concept (e.g., a suspicious address is a justification for fraud). 
For each ``transaction reason'', analysts were asked to rank the importance of each feature on a scale of 0-4, where 0 corresponded to the feature being unimportant and 4 corresponded to the feature being most important. 
To compute a feature alignment, we average over all of the analyst scores. For each analyst indexed by $i$, we refer to their provided importances as $\text{score}_i$ which maps a transaction $x$ and the $j$th feature of an explanation $E(x,f)_j$ to a value ranging from 0 to 4. If a transaction $x$ had multiple reasons labeled to it, we would select the reason that gave it the maximum score. We use the following formula to compute the average feature alignment (AVG FA) for a given explainer $E$:
\begin{equation}
    \text{AVG FA} (E) = \mathbb{E}_{(x,y) \sim \mathcal{D}}[\frac{1}{n |A|}\sum_{i=1}^{|A|} \sum_{j=1}^{n} \text{score}_i (x, E(x,f)_j)]
\end{equation}
where in our setting, the number of features in an explanation $n=6$, which is the number of non-zero features in the sparse explanation, and the number of analysts $|A|=3$. An explainer with a higher AVG FA value means that it uses features that align more with analyst priors.




\subsection{Feature Alignment Results}

\begin{table}[h!]
\centering
\caption{Average feature alignment (higher is better) for each explainer across fraudulent and legitimate transactions. Parentheses contain 90 percent CIs. Legitimate transactions score lower due to analysts considering all features similarly important when fraud is not detected.
}



\begin{tabular}{l | l | l  }
& Fraudulent Concepts & Legitimate Concepts  \\

\hline
TreeInterpreter &  2.21 (2.09, 2.37) & 1.28 (1.26, 1.29) \\
LIME &  2.05 (1.95, 2.15) & 1.22 (1.21, 1.23) \\
TreeSHAP & 2.15 (2.02, 2.32) & 1.22 (1.20, 1.23) 
\end{tabular}
\label{tab:qualitative}
\end{table}

As shown in Table~\ref{tab:qualitative}, we find that particularly for \emph{Fraudulent Concepts}, there is higher feature alignment, AVG FA, for both TreeInterpreter and TreeSHAP compared to LIME. This aligns with Table~\ref{tab:performancecompare} where we found that TreeInterpreter and TreeSHAP outperformed LIME for \simevals{} excluding $x$. 
One potential reason for why LIME has higher PDR and lower AVG FA is because the same feature appears more often in the LIME explanation compared to the two other explainers (as shown in Table~\ref{tab:unique}).
Given the explanations are sparsely populated, an explanation which consistently ranks one feature as being important would likely be less useful in distinguishing fraudulent from legitimate transactions. 
These trends do not hold for \emph{Legitimate Concepts}. Interviews with analysts revealed that explainers' AVG FA scores for legitimate transactions were significantly lower across the board because all features could be considered to be similarly important when the transaction is legitimate. 
The fact that analysts consider all features somewhat important for legitimate transactions might mean that any drastic dimensionality reduction may be unnatural. However, it is possible that analysts can adapt to this set-up over time. 
These results provide mixed evidence for the use of explainers as a dimensionality reduction technique, though a user study would be necessary to evaluate its benefits and drawbacks.

\section{Conclusion}

We conduct the first comparison of \simevals{} against existing user study findings for the real-world use case of decision support for fraud detection. 
We find that \simevals{} results generally agreed with findings from the user study by~\citet{amarasinghe2022importance}, which is that there was no statistical difference in predictiveness of fraud between the three explanation methods considered, despite limited statistical power due to sample size.
This finding suggests that \simevals{} could have been used to identify better choices of explainers for the use case and provides additional evidence in favor of using \simevals{} before running expensive user studies. 
Furthermore, we use \simevals{} to evaluate new hypotheses and find promising evidence in favor of using explanations as a dimensionality reduction technique.
We hope this work serves as a guideline to illustrate the potential uses of \simevals{} in real-world contexts both as a way to both verify whether candidate explanation methods are predictive of a use case as well as to explore experimental design set-ups. 
\newpage

\bibliographystyle{iclr2023_conference}
\bibliography{ref}

\newpage
\appendix

\section{Additional Figures}

\begin{table}[h!]
    \centering
    \caption{A list of different values used for each hyperparameter included in the grid search. In total, 12 different hyperparameter combinations were tested.}
    \begin{tabular}{c|c}
         Hyperparameter & Possible Values\\
         \hline
         Model Family & Random Forest / Decision Tree\\
         Minimum Samples at Leaf & 5 / 10 / 15 / 20 / 25 / 30\\
    \end{tabular}
    \label{tab:hyperparams}
\end{table}

\begin{table}[h!]
\centering
\caption{A confusion matrix of analyst and \simeval{} predictions. Each entry contains the portion of transactions which were actually fraud, followed by the number of transactions, given that particular combination of \simeval{} and analyst predictions. The \simevals{} in the comparison were trained on $(\hat{y}, x, E)$ for $E \in \{ \textit{TreeInterpreter, LIME, TreeSHAP}\}$, respectively.}

\begin{tabular}{llll}
\toprule
Analyst Decision &       approved &       declined &     suspicious \\
Simulated User Prediction &                &                &                \\
\midrule
False &  0.0650, N=354 &  0.1111, N=045 &  0.0000, N=007 \\
True  &  0.3962, N=053 &  0.6562, N=032 &  0.5714, N=007 \\
\bottomrule
\end{tabular}

\begin{tabular}{llll}
\toprule
Analyst Decision &       approved &       declined &     suspicious \\
Simulated User Decision &                &                &                \\
\midrule
False &  0.0729, N=384 &  0.1471, N=034 &  0.2857, N=007 \\
True  &  0.4286, N=042 &  0.6296, N=027 &  0.6667, N=006 \\
\bottomrule
\end{tabular}

\begin{tabular}{llll}
\toprule
Analyst Decision &       approved &       declined &     suspicious \\
Simulated User Decision &                &                &                \\
\midrule
False &  0.0877, N=365 &  0.0750, N=040 &  0.2222, N=009 \\
True  &  0.3000, N=050 &  0.6897, N=029 &  0.4286, N=007 \\
\bottomrule
\end{tabular}

\label{tab:analyst_conf_mat}
\end{table}

\begin{table}[h!]
\centering
\caption{Summary analysis which aims to explore the repetitiveness of each explainer. For each explainer, we find the variance $var_{i \in \text{features}}(p_i)$, where $p_i$ is the portion of transactions shown to the analysts which had a nonzero explanation for feature $i$. We also found the number of unique features which ever received a nonzero explanation in transactions shown to the original user study analysts. Higher variance and lower number of unique features indicates that an explainer has more uninformative features.}
\begin{tabular}{l | l | l  }
& Variance & Number of Unique Features (out of 112)  \\

\hline
TreeInterpreter &  0.006 & 104 \\
LIME &  0.030 & 31 \\
TreeSHAP & 0.015 & 89 
\end{tabular}
\label{tab:unique}
\end{table}

\end{document}